\pdfoutput=1

\documentclass[11pt]{article}

\usepackage{naacl2021}

\usepackage{times}
\usepackage{latexsym}

\usepackage[T1]{fontenc}

\usepackage[utf8]{inputenc}

\usepackage{microtype}

\usepackage{graphicx}
\usepackage{amsmath}
\usepackage{amssymb}
\usepackage[shortlabels]{enumitem}
\usepackage[font={small}]{caption}
\usepackage[multiple]{footmisc}
\usepackage{multirow}
\usepackage{booktabs}

\newcommand\jb[1]{\textcolor{purple}{[JB: #1]}}
\newcommand\ag[1]{\textcolor{blue}{[AG: #1]}}

\newcommand\roberta{\textsc{RoBERTa}}
\newcommand\robertas{\textsc{RoBERTa}-$512$}
\newcommand\robertal{\textsc{RoBERTa}-$4096$}
\newcommand\softmax{$\mathrm{softmax}$}
\newcommand\comment[1]{}

\newcommand\balpha{\boldsymbol{\alpha}}
\newcommand\bbeta{\boldsymbol{\beta}}

\newcommand\inner[2]{\langle#1, #2\rangle}

\usepackage{float}     

\title{Value-aware Approximate Attention}

\author{
  Ankit Gupta \\
  Tel Aviv University \\
  {\tt {\normalsize ankitgupta.iitkanpur@gmail.com}} \\\And
  Jonathan Berant \\
  Tel Aviv University, \\
  Allen Institute for AI \\
  {\tt {\normalsize joberant@cs.tau.ac.il}} \\}

\begin{document}
\maketitle

\setlength{\abovedisplayskip}{6.5pt}
\setlength{\belowdisplayskip}{6.5pt}

\begin{abstract}
Following the success of dot-product attention in Transformers, numerous approximations have been recently proposed to address its quadratic complexity with respect to the input length. However, all approximations thus far have ignored the contribution of the \emph{value vectors} to the quality of approximation. In this work, we argue that research efforts should be directed towards approximating the true output of the attention sub-layer, which includes the value vectors. We propose a value-aware objective, and show theoretically and empirically that an optimal approximation of a value-aware objective substantially outperforms an optimal approximation that ignores values, in the context of language modeling. Moreover, we show that the choice of kernel function for computing attention similarity can substantially affect the quality of sparse approximations, where kernel functions that are less skewed are more affected by the value vectors. 
\end{abstract}

\section{Introduction}\label{sec:intro}

The Transformer architecture \cite{vaswani2017attention} has been widely successful in a wide range of natural language processing tasks, including machine translation \cite{edunov2018understanding}, language modeling \cite{roy2020efficient}, question-answering \cite{karpukhin2020dense}, and many more. Transformers pre-trained on large amounts of text with a language modeling (LM) objective, have become the standard in NLP, exhibiting surprising amounts of linguistic and world knowledge \cite{peters2018elmo, devlin2018bert, petroni2019language, hewitt2019structural,Roberts2020t5kb}.

The contextualizing component of the Transformer is the attention layer where all positions in an input sequence of length $L$ aggregate information from the entire sequence in parallel. At its core, given $L$ query, key and value vectors, the \textit{dot-product attention} function outputs\footnote{Usually, the term is $\mathrm{softmax}(QK^\top / \sqrt{d})V$ but $\sqrt{d}$ can be dropped via scaling of queries and keys.} $\mathrm{softmax}(QK^\top)V$ where the \softmax{} function is applied row-wise on the matrix $QK^\top \in \mathbb{R}^{L \times L}$,
consisting of similarity scores of the query-key pairs. Unfortunately, computing $\Omega(L\cdot L)$ similarity scores is prohibitive for long sequences. 

To alleviate this, past work proposed to compute an approximation 
of $\mathrm{softmax}(QK^\top)$. One major line of research focused on \textit{sparse attention} variants, where only a few similarity scores are computed per position, and the rest are ignored. Methods differ by which query-key pairs are selected \cite{child2019generating, ye2019bp, qiu2019blockwise, roy2020efficient, kitaev2020reformer, beltagy2020longformer,gupta2020gmat}. 
A second line of research explored \textit{dense} variants  \cite{katharopoulos2020transformers,Wang2020LinformerSW,tay2020sparse} (cf.\ \cite{tay2020efficient} for a survey). E.g., instead of computing the attention scores exactly for only a few query-key pairs, \cite{Choromanski2020RethinkingAW} compute an approximation of scores for all pairs.

In this work, we point to a lacuna in current research on efficient Transformers. While recent work focused on approximating the attention scores $\mathrm{softmax}(QK^\top)$, the true target of approximation should be the output of the attention sub-layer, namely $H = \mathrm{softmax}(QK^\top)V$, which also includes the value vectors, $V$. We show that ignoring value vectors leads to unwarranted consequences both theoretically and empirically.

To demonstrate the importance of value-aware approximation, we analyze \emph{optimal sparse attention}, that is, the case where, in hindsight, the model computes dot product similarity only with the most similar key vectors, while still ignoring the value vectors.
We show that in the popular masked language modeling (MLM) setup, optimal sparse attention  dramatically \emph{under-performs} compared to an optimal approximation of the true output of the attention sub-layer, $H$, leading to an error increase of $8$-$20$ points. Next, by theoretically focusing on the case where queries compute similarity to the \emph{single} most similar key vector, we show that approximating $\mathrm{softmax}(QK^\top)$ is equivalent to approximating $H$ when the value vectors $V$ satisfy strong orthogonality and norm constraints. Conversely, when they do not, ignoring $V$ can lead unbounded approximation error. 

Second, we discuss the kernel-based view of attention, where efficiency is gained by replacing the exponential kernel (corresponding to $\mathrm{softmax}$) with other kernel functions \cite{katharopoulos2020transformers}. We theoretically show that while in the exponential kernel case (corresponding to $\mathrm{softmax}$), the effect of the norm of the value vectors is potentially small, switching to other kernels can dramatically increase the importance of the value vectors. We empirically test this by comparing optimal sparse attention
given different kernel functions, and see that indeed approximation quality decreases when replacing the exponential kernel, 

To conclude, we theoretically and empirically show that approximating the attention score matrix alone is insufficient, and propose that the research community should instead approximate the true output of the sub-attention layer, which importantly includes value vectors. Our code and trained models are available at \url{https://github.com/ag1988/value_aware_attn}.

\section{Background}\label{sec:method}


We review the kernel-based view of attention \cite{tsai2019transformer}, which will be instructive in \S\ref{sec:optimal}.

\paragraph{Generalized Attention} 
Let $\kappa(x,y) = \inner{\Phi(x)}{\Phi(y)} \geq 0$ be a kernel function with feature map $\Phi:\mathbb{R}^d \mapsto \mathcal{H}$ for some implicit reproducing kernel Hilbert space (RKHS) $\mathcal{H}$. Given a query vector $q$, keys $k_1,\ldots,k_L$, values $v_1,\ldots,v_L$, all in $\mathbb{R}^d$:
\begin{equation}\label{eqn:attention}
\mathrm{att}_\kappa(q,k_1,\ldots,v_1,\ldots) = \frac{\sum_{i=1}^{L} \kappa(q,k_i)v_i}{\sum_{i=1}^{L} \kappa(q,k_i)},
\end{equation}
where the normalization induces a probability distribution $\balpha$ over the value vectors with $\alpha_i = \kappa(q,k_i) / \sum_i \kappa(q,k_i)$. The most popular use case is the exponential kernel $\kappa(x,y) = \exp(\inner{x}{y})$, referred to as dot-product attention in Transformers. 
Some other examples include the degree-$2$ \textit{polynomial} kernel $\kappa(x,y) = \inner{x}{y}^2$ and the recently proposed \textit{elu} kernel $\inner{\Phi(x)}{\Phi(y)}$ with $\Phi(\cdot) = 1 + \mathrm{ELU}(\cdot)$ \cite{katharopoulos2020transformers}.

Given $L \gg d$ queries, the attention function (Eq. \ref{eqn:attention}) requires computing $L\cdot L$ similarity scores for the query-key pairs, which is prohibitive for long sequences. 
\textit{Sparse attention} variants relax this requirement and compute only a few similarity scores, ignoring the rest:
\begin{equation}\label{eqn:sparse_attention}
\mathrm{att}_{\kappa, S} = \frac{\sum_{i\in S} \kappa(q,k_i)v_i}{\sum_{i\in S} \kappa(q,k_i)},
\end{equation}
for some $S \subseteq \{1,\ldots,L\}, |S|\ll L$. 
Methods differ in how $S$ is determined given the queries and keys, and include use of locality bias \cite{beltagy2020longformer}, global memory \cite{gupta2020gmat}, and LSH hashing \cite{kitaev2020reformer}, among others.
Conversely, instead of exactly computing the attention scores only on a few query-key pairs, \textit{dense} variants compute an approximation of the true kernel values for all pairs. Such methods output $\sum_i \beta_i\cdot v_i$ for some approximation $\bbeta$ of the the true attention distribution $\balpha$ \cite{Choromanski2020RethinkingAW,peng2021random}. 

\section{Optimal Sparse Attention}\label{sec:optimal}


Prior methods for approximating attention, ignored the contribution of the values vectors $V$. As the true output of the attention sub-layer also depends on $V$, a natural question is whether it is possible to design better approximation methods by incorporating $V$, and if so, how much improvement is even possible? 

To answer this, we focus on sparse attention, and analyze the difference between an oracle sparse approximation that considers the value vectors, and an oracle approximation that does not. That is, we look at the difference between the two approximations from the perspective of \emph{expressivity}, ignoring any memory and computational constraints. We denote an optimal value-aware approximation that uses $r$ key vectors per query by \emph{optimal-v-aware-r}, and an optimal approximation that ignores value vectors by \emph{optimal-v-oblivious-r}. 
We define \emph{optimal-v-oblivious-r} as the output of Eq.~\ref{eqn:sparse_attention} in which $S$ is selected to be the $r$ indices with the highest attention scores $\alpha_i$'s. This is a natural baseline since this is what current sparse methods are trying to emulate.
We now explicitly derive and analyze the value-aware objective.


\paragraph{Value-aware objective} Let $o = \sum_{i=1}^L \alpha_iv_i$ be a convex combination of $v_1,\ldots,v_L \in \mathbb{R}^d$, corresponding to the true output of the attention sub-layer.
Let $C_r = \{\sum_{i=1}^L \beta_iv_i: \forall i\  \beta_i \geq 0, \sum_i \beta_i=1, |\{\beta_i: \beta_i > 0\}| \leq r\}$ denote the set of points in the polytope of $v_i$'s that can be expressed as a convex combination of at most $r$ value vectors $v_i$.
The goal of value-aware approximation is to solve for the point in the constrained region $C_r$ closest to the true output $o$, i.e. $\mathrm{argmin}_{\tilde{o} \in C_r} ||o-\tilde{o}||^2$. As mentioned,  this solution is termed \textit{optimal-v-aware-r}.

We consider two extreme cases of $r$: $r=1$ and $r\geq d+1$. For $r\geq d+1$, the Carath{\'e}odory Theorem \cite{Caratheodory} states that $o = \sum_i \alpha_iv_i$ can be expressed as a convex combination of at most $d+1$ $v_i$'s. Hence, if $r \geq d+1$ then $o \in C_r$ and the optimal approximation error is $0$. 
In most popular architectures, such as BERT \cite{devlin2018bert}, $d=64 \ll L$. This means that from the point of expressivity, \emph{optimal-v-aware-65} can obtain a perfect approximation. Conversely, we will show in \S\ref{sec:experiments} that the performance of \emph{optimal-v-oblivious-65} is substantially lower.


At the other extreme, when $r=1$ (a single value vector), the above objective is equivalent to $\mathrm{argmin}_{i \in (1,\dots,L)} ||o-v_i||^2$ and can be simplified as
\setlength{\abovedisplayskip}{2pt}
\setlength{\belowdisplayskip}{2pt}
\begin{equation}\label{eqn:top_1}
\begin{split}
 &\ \mathrm{argmin}_{i} ||o||^2 + ||v_i||^2 -2\inner{v_i}{o} \\
= &\ \mathrm{argmin}_{i} ||v_i||^2 -2\inner{v_i}{\sum_{j} \alpha_jv_j} \\ 
= &\ \mathrm{argmin}_{i} ||v_i||^2(0.5-\alpha_i) -\sum_{j\neq i} \alpha_j\inner{v_i}{v_j}.
\end{split}
\end{equation}

This equation induces a ranking over value vectors that \emph{depends} on the value vectors themselves, in contrast to a value-oblivious ranking induced solely by attention weights $\balpha$. 
\setlength{\abovedisplayskip}{6pt}
\setlength{\belowdisplayskip}{6pt}

If $v_1,\ldots,v_L$ are orthogonal, the above equation further simplifies to $\mathrm{argmin}_{i} ||v_i||^2(0.5-\alpha_i) - \sum_{j\neq i} \alpha_j\cdot0 = \mathrm{argmin}_{i} ||v_i||^2(0.5-\alpha_i)$. In this case, if some $\alpha_i \geq 0.5$ or if $v_1,\ldots,v_L$ have equal norms, this would further simplify to $\mathrm{argmax}_{i} \alpha_i$, and would therefore be independent of the value-vectors $v_i$'s, implying that a value-oblivious approximation would work well.

But such assumptions on $v_1, \ldots, v_L$ do not hold in general and thus an approximation that only depends on $\alpha_i$'s can be sub-optimal. E.g., let $v_1, v_2, v_3$ be orthogonal vectors $(1,0,0)$, $(0,2,0)$, $(0,0,3)$ respectively and let $\alpha_1, \alpha_2, \alpha_3$ be $0.25, 0.35, 0.4$. Then $v_3$ with the highest attention weight $\alpha_3$ has a squared distance of $3.79$ from the true output $\sum_i \alpha_iv_i$ whereas $v_1$ with the least attention weight $\alpha_1$ has only $2.49$. In this case, \emph{optimal-v-aware-1} induces exactly the opposite ranking of value vectors compared to \emph{optimal-v-oblivious-1}. Moreover, if we increase the value $3$ in $v_3$ to infinity, the approximation error will also infinitely grow. This example and, in general, Eq.~\ref{eqn:top_1} also show that the optimal ranking can be significantly different from the one induced by $\alpha_i ||v_i||$ proposed recently by \cite{kobayashi-etal-2020-attention} for obtaining better interpretability of attention models.




\paragraph{Effect of kernel function} Recently, Linear Transformer \cite{katharopoulos2020transformers} proposed to replace the existing exponential kernel with more efficient kernels. We now show that replacing the exponential kernel with a polynomial kernel can lead to a drop in quality for current sparse approximation methods.

Intuitively, because the kernel function affects the skewness of $\balpha$, it also affects the difference between the ranking induced by the optimal-value-aware approximation and the optimal-value-oblivious one. For simplicity, consider the case of orthogonal value vectors in which Eq.~\ref{eqn:top_1} simplifies to $\mathrm{argmin}_{i} ||v_i||^2(0.5-\alpha_i)$. From Eq.~\ref{eqn:attention}, we have $\alpha_i = \kappa(q,k_i) / \sum_j \kappa(q,k_j)$ which is $\inner{q}{k_i}^C / \sum_j \inner{q}{k_j}^C$ for the degree-$C$ polynomial kernel. For $C = 0$, we have $\alpha_i = 1/L$, which gives $\mathrm{argmin}_{i} ||v_i||^2$. In this case, the value vectors become crucial when $\balpha$ is uniform. On the other hand, assuming distinct inner products, for $C \gg 0$ we will obtain $\max_i \alpha_i \geq 0.5$, thereby reducing us to $\mathrm{argmax}_{i} \alpha_i$, where value vectors do not affect the approximation. The complexity of the Transformer grows exponentially with the degree $C$ and thus in practice a low $C$ must be used (e.g., degree-$2$ polynomial).
In such case, $\balpha$ is likely to be less skewed compared to the exponential kernel and more likely to induce a sub-optimal ranking.

In the next section, we empirically verify the above observations and show a significant performance gap between value-oblivious approximations and value-aware ones.


\section{Experiments}\label{sec:experiments}

We empirically verify our observations in the context of training causal and masked language models, which are known to strongly correlate with performance on downstream applications \cite{radford2019language,devlin2018bert}. 

\paragraph{Masked LM task} We form examples by sampling sequences and replacing sub-words with \texttt{<mask>} following the procedure in \cite{devlin2018bert}. The model is trained to maximize the log probability of the masked out tokens and we evaluate the \emph{error} of the model as the percentage of masked tokens predicted incorrectly. As approximate attention becomes increasingly relevant for long sequences, we train \robertal{} on sequences of length $4096$ (Fig.~\ref{figure:mlm_training}). Training was warm-started using \roberta{}-base \cite{liu2019roberta}. Full details on the experimental setup are in \S\ref{sec:mlm_data}. After training the model for $\sim2.5$M steps, the error of the model (that is, proportion of incorrect predictions) on the evaluation set was $24.2$ (compared to $26.6$ for an analogous training on $512$-long sequences), ensuring that tokens in \robertal{} indeed attend over longer distances and result in higher quality representations. We then replace the attention function of the trained model with various approximation schemes and evaluate the resulting model on the evaluation set. 

\begin{figure}[h!]
    \centering
    \includegraphics[scale=0.45]{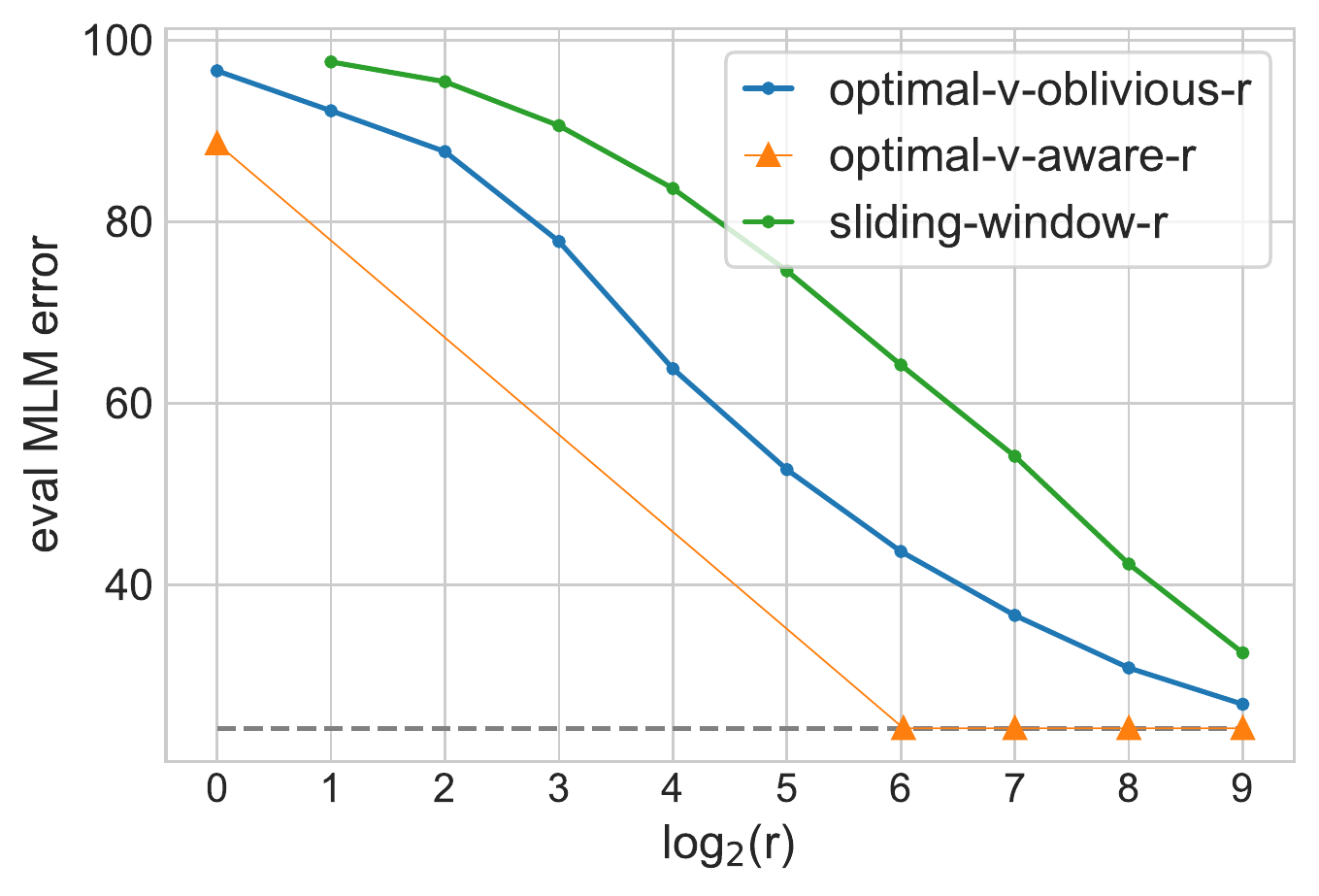}
    \caption{Evaluation of MLM error of \robertal{} after replacing vanilla attention with approximation schemes. Dashed line denotes error using vanilla attention.}
\label{figure:mlm_eval_top_r}
\end{figure}

We first compare \emph{optimal-v-oblivious-r} to \emph{optimal-v-aware-r}. We know that the approximation error of value-aware approximation is $0$ for $r > 64$. For $r=1$, we exhaustively go through all possible values and choose the one that minimizes the value-aware objective. As seen in Fig.~\ref{figure:mlm_eval_top_r} and Table~\ref{table:mlm_error}, 
there is substantial gap between the two approximations. For instance, \emph{optimal-v-oblivious-65} gives an MLM error of $43.5$ whereas the error of \emph{optimal-v-aware-65} is $24.2$, since it can perfectly approximate full attention. Moreover, we compare \emph{optimal-v-oblivious-r} to existing approximations: (a) \emph{sliding-window-r}, where a position attends to $r/2$ positions to its left and right), (b) LSH attention \cite{kitaev2020reformer} and (c) ORF attention \cite{Choromanski2020RethinkingAW}. Fig.~\ref{figure:mlm_eval_top_r} shows that \emph{sliding-window-r} trails behind \emph{optimal-v-oblivious-r}.
LSH attention, which tries to emulate \emph{optimal-v-oblivious-r}, either requires a large number of hash rounds or a large chunk size. Similarly, the dense approximation, ORF, provides an unbiased approximation of the exponential kernel but suffers from high variance in practice.


\begin{table}[h]\setlength{\tabcolsep}{3.6pt}
    \scriptsize
    \centering
    \begin{tabular}{c|c|c|c|c|c|c|c}\hline
    exact & \begin{tabular}[c]{@{}c@{}} OVO\\ $1$\end{tabular} & \begin{tabular}[c]{@{}c@{}} OVO\\ $65$\end{tabular} & \begin{tabular}[c]{@{}c@{}} OVA\\ $1$\end{tabular} & \begin{tabular}[c]{@{}c@{}} OVA\\ $65$\end{tabular} & \begin{tabular}[c]{@{}c@{}} ORF\\ $256$ features\end{tabular} & \begin{tabular}[c]{@{}c@{}} LSH\\ $r=64$ \\ $4$-rounds \end{tabular} & \begin{tabular}[c]{@{}c@{}} LSH\\ $r=512$ \\ $16$-rounds \end{tabular} \\\hline
    24.2 & 96.6 & 43.5 & 88.6 & 24.2 & 89.79 & 90.39 & 26.11 \\\hline
    \end{tabular}
    \caption{MLM error of \robertal{} on the evaluation set using approximate attention described in \S\ref{sec:experiments}. OVO: \emph{optimal-v-oblivious}, OVA: \emph{optimal-v-aware}. In LSH, each query attends to a total of $r$ keys per hash round.}
    \label{table:mlm_error}
\end{table}



\begin{table}[h]\setlength{\tabcolsep}{5pt}
    \scriptsize
    \centering
    \begin{tabular}{c|c|c|c|c|c}\hline
                 & exact & OVO-1 & OVO-65 & OVA-1 & OVA-65 \\\hline
    exponential  & 30.5 &  1031.1 &  33.5 &  280.3 &  30.5 \\\hline
    polynomial (deg 2) & 34.2 &  6700.2 &  310.2 &  1005.4 &  34.2 \\\hline
    elu          & 35.3 &  1770.6 &  62.7 &  837.4 &  35.3 \\\hline
    \end{tabular}
    \caption{Evaluation perplexity of  models using approximate attention. OVO: \emph{optimal-v-oblivious}, OVA: \emph{optimal-v-aware}.}
    \label{table:lm_loss}
\end{table}

\begin{figure}[h!]
    \centering
    \includegraphics[scale=0.45]{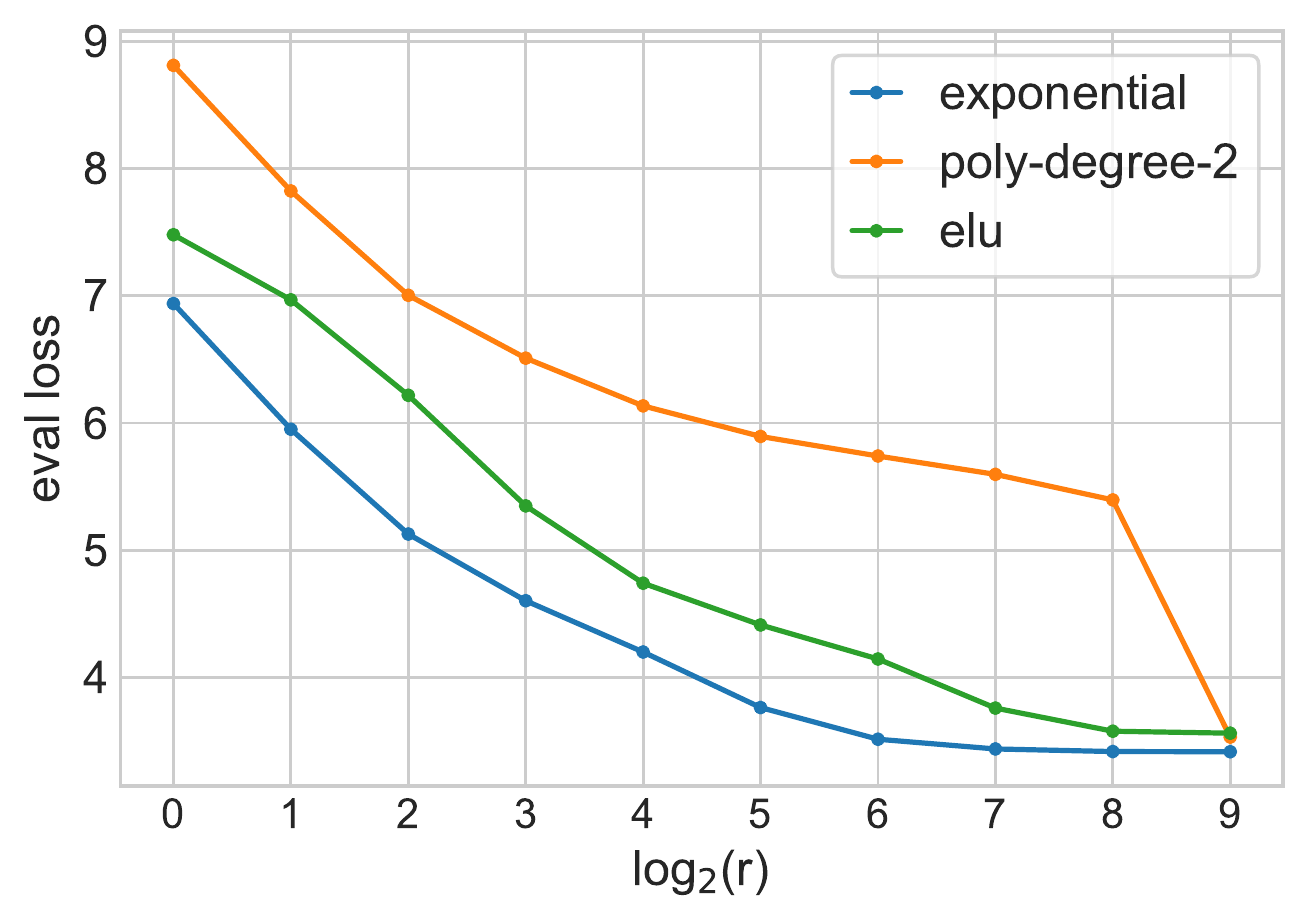}
    \caption{Evaluation loss (base e) of \emph{optimal-v-oblivious-r} oracle on the causal LM task for distinct kernel functions.}
\label{figure:kernels_top_r}
\end{figure}

\paragraph{Causal LM task} To investigate the effect of the kernel function on the quality of value-oblivious methods, we train a $6$-layer Transformer LM over 512 tokens on WikiText-103 \cite{Merity2017PointerSM} (details in \S\ref{sec:wikitext}). We train $3$ models with identical hyperparameters using the exponential, degree-$2$ polynomial, and elu kernels respectively and evaluate the trained models with value-aware and value-oblivious approximations. 
Again, \emph{optimal-v-aware-r} substantially outperforms \emph{optimal-v-oblivious-r} (Table~\ref{table:lm_loss}), pointing to the potential of working on approximating the value-aware objective. 
More importantly, comparing the approximation quality across different kernel functions (Fig.~\ref{figure:kernels_top_r}), we see that the gap between the three kernels is small when using full attention (512 keys) vectors. However, convergence is much slower for the elu kernel, and especially the degree-$2$ polynomial, demonstrating that the approximation based on the top-$r$ key vectors is sub-optimal when switching to a less skewed kernel, which is more affected by the value vectors.

\section{Conclusions}
In this work, we provide theoretical and empirical evidence against current practice of focusing on approximating the attention matrix in Transformers, while ignoring the value vectors. We propose a value-aware objective and argue that the efforts to develop more efficient Transformers should consider this objective function as a research target.

\section*{Acknowledgments}
This research was partially supported by 
The Yandex Initiative for Machine Learning, and the European Research Council (ERC) under the European Union Horizons 2020 research and innovation programme (grant ERC DELPHI 802800).

\bibliography{all}
\bibliographystyle{acl_natbib}

\clearpage

\appendix

\section{Supplemental Material}
\label{sec:supplemental}

\subsection{Masked LM task}\label{sec:mlm_data}
\begin{figure}[ht]
    \centering
    \includegraphics[scale=0.45]{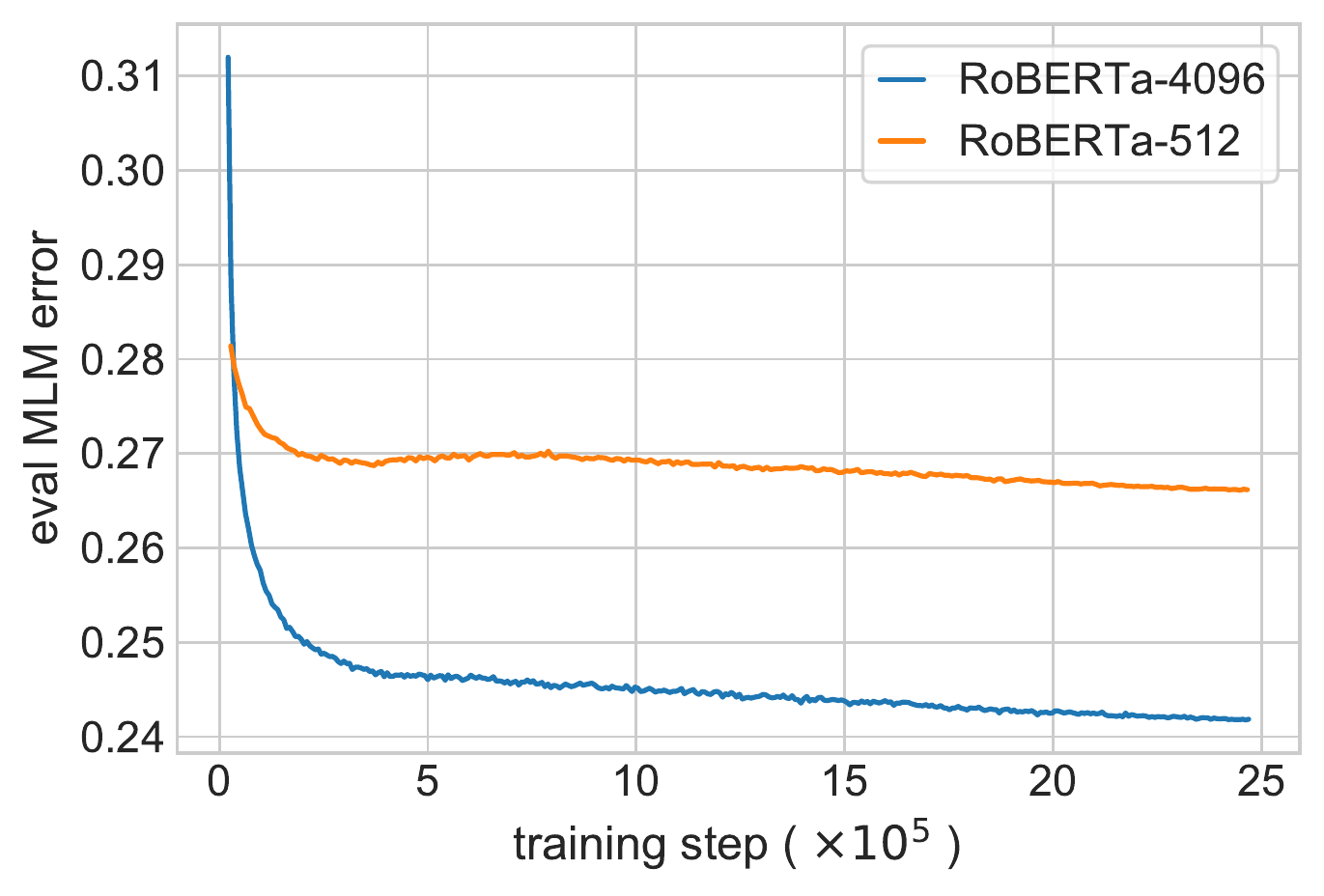}
    \caption{Evaluation error on the MLM task using vanilla attention (computing the full attention matrix).}
\label{figure:mlm_training}
\end{figure}

The instances for the MLM task (\S\ref{sec:experiments}) were formed separately using the corpora listed in Table  ~\ref{table:mlm_data}. For each dataset, after appending \texttt{</s>} token at the end of each document, the documents were arranged in a random order and concatenated into a single long text which was then tokenized into a list of sub-words. Depending upon the final input sequence length $L$ of the experiment ($512$/$4096$) this list was chunked into full length $L-2$ sequences which were then masked randomly following \cite{devlin2018bert} and enclosed within \texttt{<s>} and \texttt{</s>} tokens. To handle sequences longer than $512$ tokens, the positional embeddings were used following \cite{gupta2020gmat}. The learning curves of \robertas{} and \robertal{} are in Fig.~\ref{figure:mlm_training}.

\begin{table}[h!]\setlength{\tabcolsep}{6pt} 
    \scriptsize
    \centering
    \begin{tabular}{c|c|c|c}\hline
    corpus & all & training & evaluation \\\hline
    Wikipedia ($10$/$2017$) & $2.67$B & $1.53$B & $1.02$M\\\hline
    BookCorpus \cite{Zhu_2015_ICCV} & $1.06$B & $1.02$B & $1.02$M\\\hline
    ArXiv \cite{Cohan2018ADA} & $1.78$B & $1.53$B & $1.02$M\\\hline
    PubMed \cite{Cohan2018ADA} & $0.47$B & $510$M & $1.02$M\\\hline
    PG19 \cite{raecompressive2019} & $3.06$B & $510$M & $1.02$M\\\hline
    \end{tabular}
    \caption{Number of tokens in the datasets used for MLM training.}
    \label{table:mlm_data}
\end{table}

\paragraph{Hyperparameters} For convenience, we denote the training hyperparameters using the following abbreviations, INS: number of training instances, BSZ: number of instances in a batch, ISZ: instance size, SQL: final input sequence length after rearranging BSZ instances each of length ISZ, LR: learning rate, WRM: linear LR warm-up proportion, EP: number of epochs, STP: number of optimizer steps, GAC: gradient accumulation steps, POSq: whether (y/n) $q$ part is included in positional embeddings. The hyperparameters are listed in Table~\ref{table:hyperparams_mlm}.

\begin{table}[h!]\setlength{\tabcolsep}{2pt}
    \scriptsize
    \centering
    \begin{tabular}{c|c|c|c|c|c|c|c|c|c} \hline
        model & init & BSZ & ISZ & SQL & LR & WRM & EP & STP & POSq \\\hline
        \robertas{} & \roberta{} & $8$ & $512$ & $512$ & $5$e-$6$ & $0.1$ & $2$ & $2.476$M & n \\\hline
        \robertal{} & \roberta{} & 8 & $512$ & $4096$ & $5$e-$6$ & $0.1$ & $2$ & $2.476$M & y \\\hline
    \end{tabular}
    \caption{Training hyperparameters. Common parameters: INS=$10$M, dropout-rate=$0.0$, optimizer=Bert-Adam, $\beta_1$=$0.9$, $\beta_2$=$0.98$, weight-decay=$0.01$, max-grad-norm=$5.0$, seed=$42$, GAC=$1$.}
    \label{table:hyperparams_mlm}
\end{table}

\paragraph{Details of LSH attention} Given $L$ queries and $L$ keys in $\mathbb{R}^d$, in each hash round, we sample a new matrix $R \in \mathbb{R}^{\frac{C}{2} \times d}$ of standard gaussians and hash the queries and keys as $H_R(x) = \mathrm{argmax}([-Rx;Rx]) \in \{1,\ldots,C\}$. We rearrange the queries (and similarly keys) according to their hash value, breaking ties using the original position, and then chunk them into $L/B$ chunks of $B$ vectors each. Denoting these chunks as $Q_1,\ldots,Q_{L/B}$ and $K_1,\ldots,K_{L/B}$, for each query in $Q_i$ we compute its similarity scores with respect to all keys in $K_{i-1},K_i$. I.e.~in each hash round a query attends to $r=2B$ keys. For each query, these similarity scores are accumulated over different hash rounds, and at the end normalized by their sum to get normalized attention scores over the keys. As recommended in the original paper \cite{kitaev2020reformer}, we use $C=2L/B=4L/r$ which in practice can be sub-optimal as rearrangement destroys the original locality structure. 

\paragraph{Details of ORF attention} Given $L$ queries and $L$ keys in $\mathbb{R}^d$ we divide each vector by $d^{\frac{1}{4}}$ to account for the temperature term in dot-product attention. For a given number $F$ of features, we sample a random orthogonal matrix $R \in \mathbb{R}^{F \times d}$ as described in \cite{saxe2013exact} and provided as a tensor initialization option in PyTorch. We then map each vector to the feature space as $\Phi(x) = \frac{1}{\sqrt{F}}\exp\left(Rx - \frac{||x||^2}{2}\right)\in \mathbb{R}^F$ where ($-$) and $\exp$ operations are applied element-wise. Similarity score of a query-key pair $(q, k)$ is computed as $\inner{\Phi(q)}{\Phi(k)}$ and and is normalized by the sum of the similarity scores of $q$ with all the keys. Computing this directly leads to numerical instability so we instead compute $\Phi(q) = \frac{1}{\sqrt{F}}\exp\left(Rq - \frac{||q||^2}{2} - \max(Rq)\right)$ for queries and $\Phi(k) = \frac{1}{\sqrt{F}}\exp\left(Rk - \frac{||k||^2}{2} - \max(RK)\right)$ where $K$ is the matrix of all keys and $\max$ is over all elements of input. 

The main idea behind ORF attention is that, for a vector $w$ of standard gaussians, $\inner{w}{x} \sim \mathcal{N}(0,||x||^2)$ and from the properties of log-normal distributions, $\mathbb{E}_{w}[\exp(\inner{w}{x})] = \exp(\frac{||x||^2}{2})$. So, $\mathbb{E}_{w}[\exp(\inner{w}{q})\cdot\exp(\inner{w}{k})] = \mathbb{E}_{w}[\exp(\inner{w}{q+k})] = \exp(\frac{||q+k||^2}{2}) = \exp(\inner{q}{k}+\frac{||q||^2}{2}+\frac{||k||^2}{2})$. Appropriately scaling both sides gives, $\mathbb{E}_{w}[\exp(\inner{w}{q} - \frac{||q||^2}{2})\cdot\exp(\inner{w}{k}-\frac{||k||^2}{2})] = \exp(\inner{q}{k})$, which is exactly the term for the exponential kernel.

\subsection{Causal LM task}\label{sec:wikitext}
For this task, we used the language modeling framework provided by Faiseq\footnote{\url{https://github.com/pytorch/fairseq}}.
\paragraph{Model and training details} number of decoder layers: $6$, hidden size: $512$, head size: $64$, number of model parameters: $156$M, dataset: WikiText-$103$, training examples: $1801350$, input sequence length: $512$, $\beta_1$=$0.9$, $\beta_2$=$0.98$, weight-decay: $0.01$, gradient clip-norm: none, learning rate: $0.0005$, learning rate schedule: inverse square root, number of warmup updates: $4000$, batch size: $128$, epochs: $20$, number of steps: $31520$, minimum context-window during evaluation on test-set: $400$.

\end{document}